\documentclass{bmvc2k}

\usepackage{amssymb}

\title{SVL-Adapter: Self-Supervised Adapter for Vision-Language Pretrained Models}

\addauthor{Omiros Pantazis}{omiros.pantazis.16@ucl.ac.uk}{1}
\addauthor{Gabriel Brostow}{g.brostow@cs.ucl.ac.uk}{1,4}
\addauthor{Kate E. Jones}{kate.e.jones@ucl.ac.uk}{1}
\addauthor{Oisin Mac Aodha}{oisin.macaodha@ed.ac.uk}{2,3}

\addinstitution{
University College London
}
\addinstitution{
University of Edinburgh
}

\addinstitution{
Alan Turing Institute
}

\addinstitution{
Niantic
}

\runninghead{Pantazis, Brostow, Jones, and Mac Aodha}{SVL-Adapter}

\def\eg{\emph{e.g}\bmvaOneDot}

\def\ie{\emph{i.e}\bmvaOneDot}
\def\etc{\emph{etc.}\bmvaOneDot}

\begin{document}

\maketitle

\begin{abstract}
Vision-language models such as CLIP are pretrained on large volumes of internet sourced image and text pairs, and have been shown to \emph{sometimes} exhibit impressive zero- and low-shot image classification performance. However, due to their size, fine-tuning these models on new datasets can be prohibitively expensive, both in terms of the supervision and
compute required. 
To combat this, a series of light-weight adaptation methods have been proposed to efficiently adapt such models when limited supervision is available. 
In this work, we show that while effective on internet-style datasets, even those remedies under-deliver on classification tasks with images that differ significantly from those commonly found online. 
To address this issue, we present a new approach called SVL-Adapter that combines the complementary strengths of both vision-language pretraining and self-supervised representation learning. 
We report an average classification accuracy improvement of 10\% in the low-shot setting when compared to existing methods, on a set of challenging visual classification tasks. 
Further, we present a fully automatic way of selecting an important blending hyperparameter for our model that does not require any held-out labeled validation data. Code for our project is available here: \url{https://github.com/omipan/svl_adapter}.
\end{abstract}

\section{Introduction}

Learning transferable representations of visual data is a core problem in computer vision. 
Until recently, the standard approach for tasks like supervised image classification involved training models to predict the discrete class labels depicted in a collection of images. 
However, we have begun to see a new set of approaches that make use of large-scale image and natural language text pairs collected from the internet as a source of training signal. 
Methods such as CLIP~\cite{radford2021learning} and ALIGN~\cite{jia2021scaling} pretrain on hundreds of millions of text and image pairs, from which they exhibit impressive transfer in both the zero- and low-shot settings~\cite{hu2022pushing}. 

\begin{figure}[t]
\centering
\includegraphics[width=1.0\textwidth]{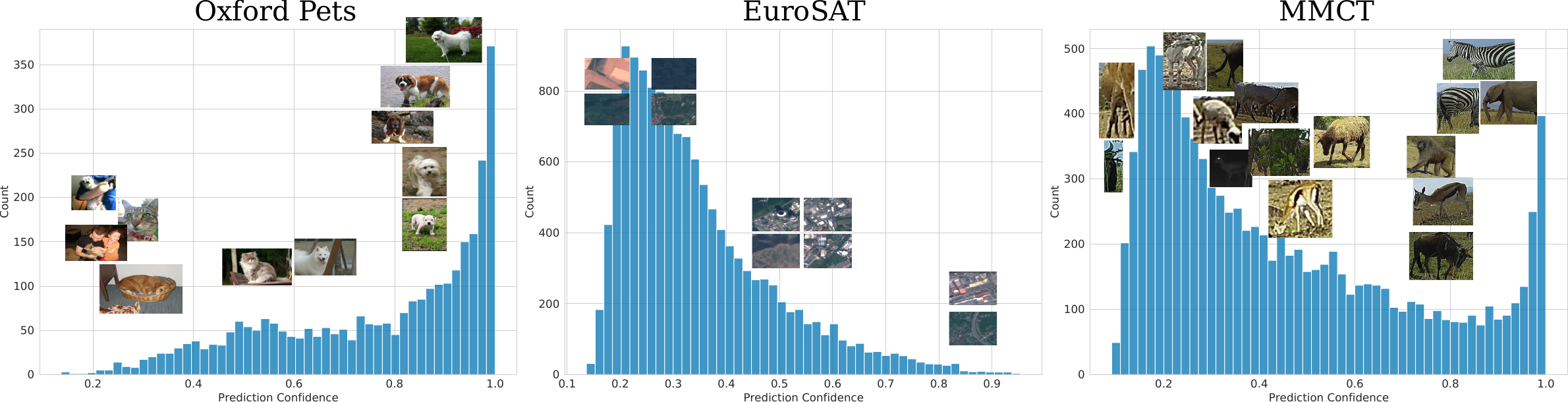}
\vspace{-15pt}
\caption{Histograms of confidence scores from zero-shot predictions of CLIP~\cite{radford2021learning}. 
Representative images for different confidence bins are also displayed.  
(Left) CLIP is confident for images from datasets representative of the types commonly found online. 
(Middle) However, when the data distribution is significantly different, \eg satellite images, it tends to have low confidence. 
(Right) In practice, real-world datasets can contain a mix of `easy' and `hard' images, \eg from camera traps. 
}
\label{fig:data_dist}
\end{figure}

The challenge of using models of this size is that fine-tuning them to new tasks can be prohibitively computationally expensive and potentially require significant amounts of data. 
Inspired by work in natural language processing~\cite{shin2020autoprompt,jiang2020can}, a series of subsequent  methods  have been proposed in vision to adapt these large-scale models in a data-efficient manner. 
Approaches include optimizing for the most effective text prompt~\cite{zhou2021learning,lu2022prompt,huang2022unsupervised} or refining the learned representations with compact adapter networks~\cite{gao2021clip,zhang2021tip}. 
However, the majority of these methods do not significantly change the underlying visual representations contained in models such as CLIP. 
As a result, they are fundamentally limited by the expressiveness of the original representations. 
This can be problematic if one wants to adapt these internet-trained models to new tasks that differ from the types of images commonly found on the internet (see Fig.\ref{fig:data_dist}), \eg medical image analysis~\cite{kermany2018identifying}, remote sensing~\cite{christie2018functional}, biodiversity monitoring~\cite{beery2018recognition}, \etc 
Unlike curated image collections such as ImageNet \cite{deng2009imagenet}, the aforementioned datasets exhibit more real-world challenges such as partial views, occlusion, poor illumination, diverse backgrounds, low image quality, and domain shifts. 
These properties make it difficult to perform transfer learning from models trained on conventionally curated content.

Our focus is on making CLIP-like models more effective on challenging tasks that potentially fall out of the distribution of the visual or language content that they were originally trained on. 
First, we show that existing methods that tune prompts or visual features, \eg  \cite{zhou2021learning,lu2022prompt,gao2021clip,zhang2021tip}, perform poorly on these types of challenging datasets. 
Next, we observe that recent advances in Self-Supervised Learning (SSL), \eg \cite{chen2020simple,He_2020_CVPR,grill2020bootstrap}, provide a complementary approach for learning visual representations in an self-supervised manner that can be combined with the outputs of models such as CLIP. 
This is especially relevant in practical applications where images are typically available, but supervision is lacking. 
Building on these observations, and inspired by existing adapter methods~\cite{gao2021clip}, we propose a new method for visual classification in the low-shot regime and validate it across multiple challenging visual classification tasks. 
Our approach also addresses a significant limitation of many existing adapter methods, which is their requirement for held-out labeled validation data for hyperparameter selection. 
We present an automated hyperparameter selection method that foregoes the need for labeled data by making use of model predictions directly.

Our contributions are summarized as follows: 
1) We present a new visual classification approach, SVL-Adapter, that combines the best of both large-scale vision-language pretraining and targeted self-supervised learning. 
2) We outline a method for selecting a key hyperparameter for our model that does not require obtaining any expensive held-out labeled validation data as is commonly done in existing related works. 
3) Through detailed experimental evaluation on ten conventional and six challenging visual classification tasks, we show that our SVL-Adapter outperforms existing methods on average for both zero- and low-shot learning, and is significantly better on challenging datasets whose visual properties differ from those commonly found online.

\section{Related Work}
\vspace{-5pt}
\subsection{Adapting Vision-Language Models}
Large pretrained \emph{language} models have demonstrated remarkable success across a variety of natural language tasks from question-answering and sentence completion to language translation~\cite{devlin2018bert,radford2019language,brown2020language}. %
These models are powerful, but are consequently expensive to train or fine-tune owing to their large size. 
Thus, there is a need for efficient adaptation methods to enable transfer to new datasets. 
Multiple works have sought remedies by devising lightweight adapter modules containing a small number of trainable parameters~\cite{houlsby2019parameter,stickland2019bert,hu2021lora}, bias mitigation calibration~\cite{zhao2021calibrate}, or by learning task-specific soft text prompts~\cite{lester2021power}. 
As a result, practitioners can harness the power and generality of large pretrained language models on some downstream tasks of interest.

Recently, a related line of work has been explored in the \emph{vision-language} domain through a series of approaches that take advantage of the large number of images with corresponding text descriptions that can be found online~\cite{radford2021learning,jia2021scaling,yuan2021florence,yu2022coca}.  
For example, CLIP~\cite{radford2021learning} utilizes contrastive learning between the embeddings of image and text pairs, which are encoded via separate image and language encoders. 
Essentially, CLIP learns to represent the two modalities by pulling together the image representations with their paired text ones, while repelling the text embeddings corresponding to different images. 
The advantage of these approaches is that they can learn from the large unstructured collections of images and text which are readily available on the internet~\cite{jia2021scaling}. 

Vision-language models trained on large datasets can be applied to a wide range of downstream tasks in either the zero- or low-shot setting. %
In the zero-shot classification setting, the practitioner must select the set of relevant text descriptions corresponding to the classes of interest. 
\cite{radford2021learning} showed that how these text prompts are constructed can have a significant impact on the downstream performance of models such as CLIP. 
Drawing inspiration from language model adaptation, two main families of approaches for adapting CLIP-like models to downstream visual tasks have been proposed: prompt learning and feature adaptation. 
Both these methods typically rely on the existence of a small number of labeled examples for each class in the given target dataset, \ie the low-shot setting.

Prompt learning seeks to address the major limitation of having to hand-craft the text prompts, by automating how they are constructed. 
CoOp~\cite{zhou2021learning} optimizes the prompt context using a set of learnable vectors in either a unified or class specific way. 
CoCoOp~\cite{zhou2022conditional} extended this, addressing the difficult problem of generalizing to unseen classes by learning to generate vectors conditioned on each image. 
In contrast, feature adaptation approaches directly tune the representations that are  extracted from the visual and text encoders of models like CLIP.
CLIP-Adapter~\cite{gao2021clip} added a lightweight fully connected neural network \emph{adapter} that is applied to frozen CLIP features and performs fine-tuning of its parameters with limited supervision on the downstream task of interest. 
Performance improvements are reported across a variety of datasets, with the best performing variant only adapting the visual encoder features. 
Tip-Adapter~\cite{zhang2021tip} obtains even better results by constructing a key-value cache model from the low-shot samples, and fine-tunes for a smaller number of epochs. 
A tuning-free version of Tip-Adapter was also proposed, which is faster at adapting at training time, but performs worse. 

While relatively efficient, these adapter style approaches are incapable of making large-scale changes to the underlying representations extracted from the backbone visual or text encoders. 
This poses a significant issue if the images from an evaluation task of interest significantly differ from the distribution seen by the vision-language model at training time. 
In our proposed approach, we do not restrict feature extraction to only the visual encoders learned by CLIP. 
Instead, we propose to combine the impressive zero-shot performance of CLIP and features extracted via targeted self-supervised learning.

\vspace{-5pt}
\subsection{Self-Supervised Learning (SSL)}
SSL aims to learn high quality visual representations using only the signal contained in unlabeled images, \ie without the need for additional supervision provided by humans. 
Successive advances in SSL have further shrunk the gap between supervised and unsupervised representations, when evaluated on some downstream tasks~\cite{chen2020simple,caron2020unsupervised,He_2020_CVPR,grill2020bootstrap,zbontar2021barlow,dwibedi2021little}. 
One representative family of approaches is contrastive learning~\cite{gutmann2010noise,oord2018representation,wu2018unsupervised}, where the goal is to embed augmented views of a given image close in feature space, while pushing away the representations of other images in the same batch~\cite{chen2020simple} or using a memory bank~\cite{He_2020_CVPR}. 
Methods have also been explored that focus on retrieving more informative positive examples during training that exhibit more natural image variation than can be expressed by simple artificial augmentations~\cite{ayush2020geography,azizi2021big, Pantazis_2021_ICCV}. 
Other variants on the contrastive paradigm also report strong performance without the inclusion of any negative examples during training~\cite{grill2020bootstrap,chen2021exploring}. 
There are also further SSL approaches that are not limited to instance discrimination, but instead use information from nearest neighbors~\cite{dwibedi2021little}, prototype clustering~\cite{caron2020unsupervised}, and image patch reconstruction~\cite{he2022masked}. 

Most relevant to this work is the observation that SSL has been shown to be effective on multiple `real-world' tasks, \eg remote sensing~\cite{ayush2020geography}, medical image analysis~\cite{azizi2022robust}, and biodiversity monitoring~\cite{Pantazis_2021_ICCV}. 
In these application domains, it is sometimes impossible or prohibitive to even obtain large collections of unlabeled images. 
However, \cite{cole2022does} showed that self-supervised training on ImageNet~\cite{deng2009imagenet} is still highly effective even when only using $<25\%$ of the unlabeled images at training time.
In this work, we leverage these advances in SSL by learning representations on unlabeled data, which we combine with the outputs from large vision-language models, the result of which is comparable or better than either alone. 

\vspace{-5pt}
\subsection{Few-Shot Image Classification}

Being able to learn from limited examples poses a great challenge for computer vision that many few-shot learning (FSL) approaches have attempted to tackle, \eg \cite{santoro2016meta,vinyals2016matching,ravi2016optimization,finn2017model,snell2017prototypical,ren2018meta,wang2020generalizing,zhang2021shallow}. 
For example, meta-learning, a prominent solution for few-shot recognition \cite{santoro2016meta,ravi2016optimization,finn2017model,ren2018meta,zhang2021shallow}, utilizes a meta-learner to transfer learned knowledge from a support set of classes to enable it to perform few-shot classification on new classes. 
Interestingly, \cite{tian2020rethinking} showed that training a linear classifier on top of features learnt from SSL can outperform sophisticated meta-learning methods in few-shot image classification. 
Furthermore, \cite{hu2022pushing} showed that a very simple FSL method trained on vision-language model features \cite{radford2021learning} easily beats existing FSL methods. 
Thus, the success of transfer learning from models that are pre-trained on large-scale external data inspires us to focus our research on the efficient adaptation of these powerful models for few-shot, and zero-shot, learning.

\section{Method}
\subsection{Framework}
Given an image $\mathbf{x}$ as input, our goal is to predict the class label $y \in \{1, ..., K\}$ depicted in the image. 
The conventional approach would be to train an image classifier,  parameterized as a deep neural network, to perform this task. 
However, in many real world settings we typically only have access to a limited amount of training data (\ie low-shot learning), making it difficult to perform the end-to-end training of large models. 
To address this issue, recent work has shown that pretraining models with paired image and natural language supervision is an effective technique  for learning representations that can be later adapted to downstream tasks with minimal additional supervision~\cite{radford2021learning,jia2021scaling,yu2022coca}.

\begin{figure}[t]
\centering
\includegraphics[width=1.0\textwidth]{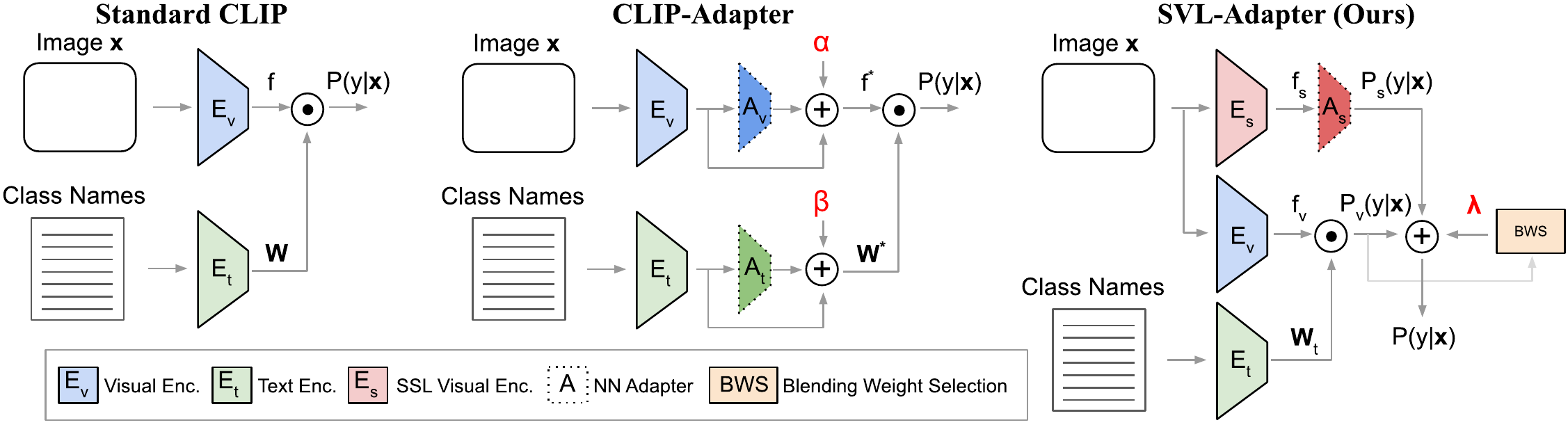}
\vspace{-15pt}
\caption{(Left) Zero-shot classification with CLIP~\cite{radford2021learning} makes use of frozen image and text encoders, $E_v$ and $E_t$, to make predictions at test time. 
(Middle) CLIP-Adapter~\cite{gao2021clip} introduces two small additional adapter neural networks, $A_v$ and $A_t$, whose outputs are combined with the visual and text embeddings from CLIP using hyperparameters $\alpha$ and $\beta$. 
(Right) Our SVL-Adapter makes use of an additional self-supervised \emph{encoder} $E_s$ which feeds into an adapter $A_s$. 
Unlike CLIP-Adapter, our approach fuses outputs at the class prediction level, where the weighting $\lambda$ of this combination is automatically controlled by our blending weight selection module. 
}
\label{fig:method}
\end{figure}

\noindent{\bf CLIP:} One such representative approach is CLIP~\cite{radford2021learning} (see Fig.~\ref{fig:method} (Left)). 
CLIP consists of two main parts: a visual encoder $E_v$ that takes an image as input and outputs a feature embedding $\mathbf{f} = E_v(\mathbf{x})$, and a text encoder $E_t$ that encodes the set of class names into the same embedding space, $\mathbf{W} = E_t(\mathcal{C})$. 
Here, $\mathbf{W} \in \mathbb{R}^{D\times K}$ is a matrix where each of the D dimensional column vectors corresponds to an embedding of the $K$ classes of interest at inference time, \ie $\mathcal{C} = \{C_1, ..., C_K\}$. 
Each entry $C_k$, is the natural language class name corresponding to the class label $y$. 
To make a prediction for an unseen test image $\mathbf{x}$, one simply performs a matrix multiplication of its feature embedding $\mathbf{f}$ and the estimated classifier weight matrix $\mathbf{W}$, \ie $P(y|\mathbf{x}) = \sigma(\mathbf{f}^{\intercal}\mathbf{W})$, where $\sigma$ is the softmax function. 
The visual encoder can be implemented as a ResNet~\cite{he2016deep} or a visual transformer~\cite{dosovitskiy2021image}, and the text encoder as a transformer~\cite{vaswani2017attention}.

\noindent{\bf CLIP-Adapter:} As presented, CLIP cannot make use of additional labeled data without performing expensive end-to-end fine-tuning or simply using the visual encoder as a fixed feature extractor. 
Several subsequent methods have been proposed to efficiently adapt large pretrained models like CLIP to the low-shot learning setting. 
One such example, is CLIP-Adapter~\cite{gao2021clip} (see Fig.~\ref{fig:method} (Middle)). 
CLIP-Adapter performs fine-tuning on light-weight residual feature adapters to improve classification performance when only modest supervision is available.  
This is achieved via an additional pair of small fully connected neural network adapters $A_v$ and $A_t$, for visual and text feature adaption respectively.
Specifically, $A_v(\mathbf{f}) = \text{ReLU}(\mathbf{f}^{\intercal}\mathbf{W}_v^1 )\mathbf{W}_v^2$ and $A_t(\mathbf{W}) = \text{ReLU}(\mathbf{W}^{\intercal}\mathbf{W}_t^1 )\mathbf{W}_t^2$.  

The output feature embedding and classifier weight matrix are then computed as
\begin{align}
\label{eqn:alpha_eqn}
\mathbf{f}^\ast =  \alpha A_v(\mathbf{f})^\intercal + (1 - \alpha)\mathbf{f}\\
\label{eqn:beta_eqn}
\mathbf{W}^\ast =  \beta A_t(\mathbf{W})^\intercal + (1 - \beta)\mathbf{W}.
\end{align}

\noindent The final prediction is then computed as $P(y|\mathbf{x}) = \sigma(\mathbf{f^\ast}^{\intercal}\mathbf{W}^\ast)$. 
During low-shot training, they need to estimate the weights, $\{\mathbf{W}_v^1, \mathbf{W}_v^2, \mathbf{W}_t^1, \mathbf{W}_t^2\}$, of the two adapters. 
This is done by applying a standard cross entropy loss on the low-shot labeled data.

\subsection{SVL-Adapter} 
One major limitation of approaches like CLIP and CLIP-Adapter is they do not make significant changes to the underlying representations encoded by the visual encoder. 
The reason for this is simple: fine-tuning large models requires lots of supervision, much more than what is available in the low-shot setting. 
This is not necessarily an issue if the images for the downstream classification task come from the same distribution as those commonly found on the internet. 
However, if the images differ significantly, \eg medical or biodiversity monitoring images, then the representations that can be extracted from the visual encoder stand a good chance of being unsuitable for the task at hand. 

We exploit the fact that while little to no label supervision may be available, we often have access to the unlabeled images at test time. 
Recent progress in self-supervised learning ~\cite{chen2020simple,He_2020_CVPR,grill2020bootstrap,he2022masked} has resulted in methods that can extract informative visual representations without requiring any supervised labels. 
To avail of these representations, we propose a new approach, the Self-supervised Vision-Language Adapter (SVL-Adapter) that combines the strengths of both vision-language pretraining and self-supervised learning (see Fig.~\ref{fig:method} (Right)). 

We introduce a new encoder $E_s$, that is trained using self-supervision on the target dataset. 
The output of this encoder is a feature vector $\mathbf{f}_s$ that is fed into an adapter network $A_s$. 
Unlike CLIP-Adapter, the output of this adapter is not a transformed feature encoding, but instead a prediction over the classes of interest, $P_s(y|\mathbf{x}) = A_s(\mathbf{f}_s) =  \sigma(\text{ReLU}(\mathbf{f}_s^{\intercal}\mathbf{W}_s^1 )\mathbf{W}_s^2)$. 
We then combine these predictions with the output of the standard Zero-shot CLIP model, $P_v(y|\mathbf{x}) = \sigma(\mathbf{f}_v^{\intercal}\mathbf{W}_t)$,  
\begin{equation}
P(y|\mathbf{x}) = \lambda P_v(y|\mathbf{x}) + (1-\lambda) P_s(y|\mathbf{x}).
\label{eqn:blend}
\end{equation}

During training we learn the weights $\{\mathbf{W}_s^1, \mathbf{W}_s^2\}$. 
We train $E_s$ on the target training dataset using a self-supervised constrastive objective~\cite{chen2020simple}, which does not require any labeled data. While training $E_s$ on large datasets could be expensive, in practice we start from an ImageNet initialized model which leads to fast convergence on relatively small downstream datasets. 
Also this step only has to be performed once as only the adapter $A_s$ needs to be retrained when the amount of supervision available changes.

\subsubsection{Blending Weight Selection} 
\label{sec:bws}
The best results for CLIP-Adapter in~\cite{gao2021clip} are obtained when the hyperparameters $\alpha$ and $\beta$ in Eqns.~\ref{eqn:alpha_eqn} and \ref{eqn:beta_eqn} are selected using held-out validation data. 
In the case of low-shot learning, any and all labeled data is precious and would likely be more valuable to use for training and not hyperparameter selection. 

To overcome this issue, we propose a conceptually simple and efficient approach for selecting our prediction blending weight $\lambda$ in Eqn.~\ref{eqn:blend} which does \emph{not} require any labeled validation data.  
From Fig.~\ref{fig:data_dist} we observe that the confidence scores corresponding to the Zero-shot CLIP predictions vary heavily among datasets. 
Based on this, and under the assumption that when CLIP is \emph{not} confident we should more heavily weigh the knowledge acquired by low-shot learning, we define $\lambda$ as being analogous to CLIP's average prediction confidence score on the $N$ test images of the given dataset, $\lambda =  \frac{1}{N}{\sum_{i=1}^N \max_{k} P(y_i = k|\mathbf{x}_i)}$.
In our experiments we compare our SVL-Adapter method to existing approaches in the setting where we select $\lambda$ using validation data (`SVL-Adapter') or where we estimate it using the outputs from CLIP as outlined above (`SVL-Adapter*').

\vspace{-5pt}
\section{Experiments}

\subsection{Implementation Details}

\noindent{\bf Datasets.} We evaluate our approach on ten standard image classifications datasets: Caltech101 \cite{fei2004learning}, OxfordPets \cite{parkhi2012cats}, StanfordCars \cite{krause20133d}, Flowers102 \cite{nilsback2008automated}, Food101 \cite{bossard2014food}, FGVCAircraft \cite{maji2013fine}, SUN397 \cite{xiao2010sun}, DTD \cite{cimpoi2014describing}, UCF101 \cite{soomro2012ucf101}, and EuroSAT \cite{helber2019eurosat}, that are typically utilized to test vision-language adaptation. Additionally, we include six challenging tasks that do \emph{not} come from the distribution of images commonly found on the internet. These datasets are: 
FMoW~\cite{christie2018functional} that contains satellite images of land or buildings, OCT~\cite{kermany2018identifying} for retina disease identification from OCT images \cite{kermany2018identifying}, and the camera trap datasets MMCT~\cite{Pantazis_2021_ICCV}, CCT20~\cite{beery2018recognition}, ICCT~\cite{islandconservation2020}, and Serengeti~\cite{swanson2015snapshot}.

\noindent{\bf Models.} To train the self-supervised feature encoder $E_s$ on each dataset for our SVL-Adapter, we use SimCLR~\cite{chen2020simple} for 200 epochs on images of size $112 \times 112$ with a standard ResNet50~\cite{he2016deep} backbone followed by a two-layer projection head. 
SSL takes on average two hours for each dataset evaluated. 
We provide additional results using alternative self-supervised methods in the supplementary material.
The trainable adapter module $A_s$ in SVL-Adapter is a two-layer neural network with 256 hidden dimensions and an output size equal to the number of classes, and is trained for 50 epochs. 
When not using our blending weight selection, $\lambda$ is tuned with a validation set, by sweeping through 20 values ranging from 0 to 1. 
For CLIP, we use the ResNet50 visual encoder $E_v$ and transformer text encoder $E_t$. 
For each of the standard datasets we use a single prompt template as defined in \cite{radford2021learning} and for the rest we define prompts that do not rely on any in-domain expertise. 
For the transformation of CLIP features we follow the pre-processing protocol of CLIP~\cite{radford2021learning}. 
We provide results for various ablations of our model in the supplementary material.  

\noindent{\bf Baselines.} To evaluate our approach, we compare with zero-shot and linear probe CLIP~\cite{radford2021learning} and state-of-the-art vision-language adaptation baselines~\cite{zhou2021learning,gao2021clip,zhang2021tip}. Specifically, we include the best performing version of CoOp~\cite{zhou2021learning} that places the class token at the end of the learnable sequence without class-specific context, the best variant of CLIP-Adapter~\cite{gao2021clip} that only adapts the visual features, and both the training-free and fine-tuned version of the more recent Tip-Adapter~\cite{zhang2021tip}. Across all scenarios, the features of the encoders stay frozen and tuning is only performed on the adapters. For low-shot learning, we construct 1, 2, 4, 8, and 16 examples per class training sets. For Zero-shot CLIP we simply apply CLIP's ResNet50 variant to the test images. 
For our Zero-shot SVL-Adapter*, training resembles the few-shot task but uses the most confident CLIP pseudolabels per predicted label instead. In our experiments we keep the 16 most confident predictions for each class. 
Evaluation always takes place on the full test set for each dataset.

\subsection{Results}

\subsubsection{Low-Shot Classification}

First we evaluate our SVL-Adapter across the multiple datasets described above in the low-shot regime with different amounts of supervision, \ie number of `shots' per class. To validate the hypothesis that images that differ from internet-style datasets need a different approach we split the datasets into two distinct sets which we term ``Standard'' and ``Challenging''. In Fig.~\ref{fig:avg_results} we observe that the top performing existing method, Tip-Adapter-F, reports less than 50\% Top-1 accuracy on average in the 16-shot setting across the ``Challenging'' tasks, while its corresponding performance on the ``Standard'' datasets is above 75\%. 
This discrepancy motivates the development of approaches that can retain the benefits of CLIP while adapting more efficiently to real-world tasks that do not exhibit internet-like visual properties.
A more detailed, per-dataset, breakdown is presented in Fig.~\ref{fig:main_results_with_pseudo}. 
We also provide these results in tabular for in the supplementary material. 

To this end, we show that SVL-Adapter, which uses self-supervised features as a starting point for visual adaptation, results in significant accuracy gains (10\% on average) across the ``Challenging'' tasks when compared to existing methods, where the largest gains are observed in the 2-shot setting and above. 
In addition, we observe that our approach still remains competitive when applied to the ``Standard'' datasets, and thus it constitutes a strong and universal baseline. 
Moreover, SVL-Adapter*, which uses our automatic blending weight selection method outlined in Sec.~\ref{sec:bws}, is comparable with SVL-Adapter on the ``Challenging'' tasks despite not requiring any labeled validation data. 
However, it is not as strong on the other datasets. 
This result is important for practitioners who wish to adapt CLIP for their tasks but cannot afford to label an additional held-out validation set for hyperparameter tuning.

\subsubsection{Zero-Shot Classification}

\begin{figure}[t]
\centering
\includegraphics[width=1.0\textwidth]{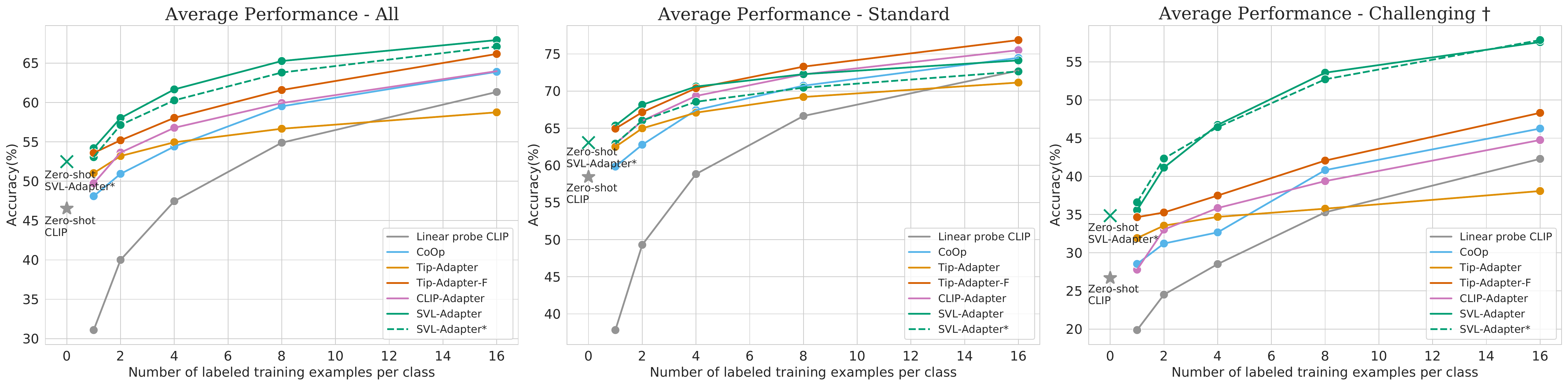}
\vspace{-15pt}
\caption{(Left) Average zero- and low-shot test Top-1 accuracy across all 16 datasets. 
(Middle) Results for the ten ``Standard'' datasets commonly used in existing work. 
Here, our SVL-Adapter is competitive with current SoTA methods. 
(Right) On the ``Challenging'' datasets, which differ more from CLIP training data, SVL-Adapter significantly outperforms existing methods. 
SVL-Adapter* refers to our approach with automatic blending weight selection as described in Sec.~\ref{sec:bws}. 
The per-dataset results are illustrated in Fig.~\ref{fig:main_results_with_pseudo}. 
}
\label{fig:avg_results}
\end{figure}

Here we attempt to further improve the already impressive zero-shot performance of vision-language models such as CLIP. 
To achieve this, we take the most confident predicted pseudolabels from CLIP for the classes of interest as the training data for SVL-Adapter, thus making it compatible for zero-shot transfer. 
Essentially, we keep the adaptation pipeline the same and just replace the ground truth labels normally used for low-shot adaptation with pseudolabels. 
Utilizing SVL-Adapter* enables us to keep the task truly zero-shot as we do not use any labeled data for hyperparameter tuning. The baseline we compare against is the standard  zero-shot version of CLIP which also uses a ResNet50 backbone. 
The results illustrated denoted as `Zero-shot SVL-Adapter*' in Figs.~\ref{fig:avg_results} and~\ref{fig:main_results_with_pseudo} show significant improvements across the majority of the datasets when compared with the standard zero-shot CLIP baseline. Specifically, for the ``Challenging'' and the ``Standard'' datasets we record an average of 8\% and 5\% improvement in Top-1 accuracy. Thus, we can infer that vision-language models, combined with adapted self-supervised features, significantly improve zero-shot classification performance.

\subsection{Limitations} 
\begin{figure}[hb]
\centering
\resizebox{1.0\linewidth}{!}
{
\includegraphics[width=1.0\textwidth]{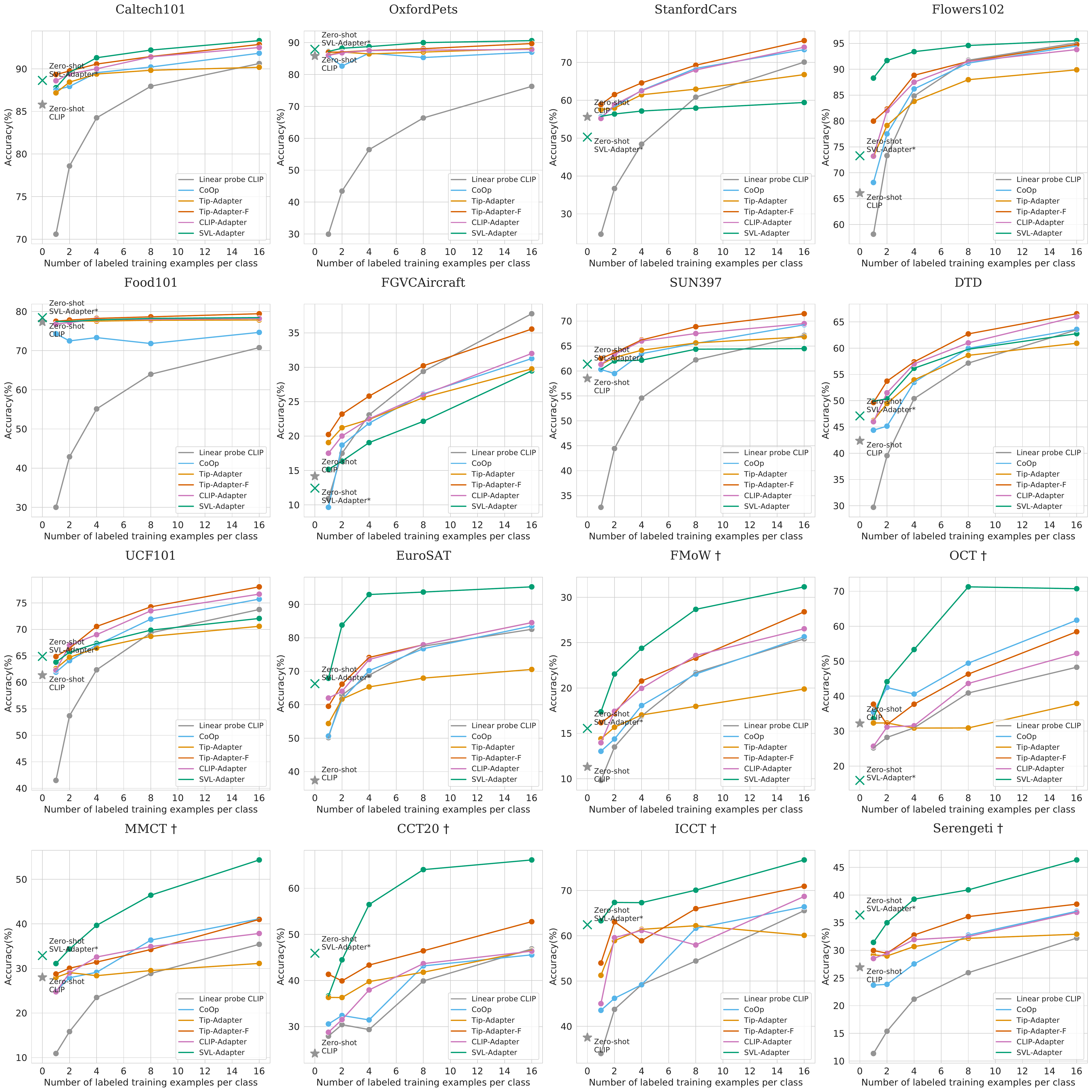}
}
\vspace{-10pt}
\caption{Per-dataset, zero- and low-shot, test Top-1 accuracy results across 16 different datasets. In each case, we report the average of three runs.
Datasets marked with $\dagger$ are added by us, and pose a significantly greater challenge to existing methods. 
A summary of the results is presented in Fig.~\ref{fig:avg_results}. 
}
\label{fig:main_results_with_pseudo}
\end{figure}

While we report large improvements for zero- and low-shot classification compared to current state-of-the-art methods on challenging datasets, there are some notable cases where we do not perform as well, \eg StanfordCars and FGVCAircraft in Fig.~\ref{fig:main_results_with_pseudo}. 
It is known that fine-grained datasets such as these pose a challenge to current SSL methods~\cite{cole2022does}. 
However, our approach is agnostic to the underlying SSL method used, and will benefit from newer, more effective representations, from this highly active research area. 
We also need to train a self-supervised representation on each dataset of interest. 
Fortunately, this is not a very time consuming operation due to the moderate size of most datasets of interest, \ie practitioners are more likely to work with `small' datasets containing thousands of images, not millions.

\section{Conclusion}

We presented SVL-Adapter, a self-supervised vision-language adapter for zero- and low-shot image classification. 
We showed that large-scale web-trained models such as CLIP fail to effectively generalize to challenging visual classification tasks that do not come from the distribution of images commonly found online. 
Furthermore, recent methods for adapting these models also fail to significantly improve performance. 
By combining the complementary strengths of self-supervised learning and vision-language pretraining, our approach results in large improvements in low-shot classification accuracy on challenging visual classification tasks without requiring any additional supervision at training time. 
We also showed that SVL-Adapter is applicable in the zero-shot learning setting, where it improves over the conventional baseline despite not requiring any additional supervision.

\vspace{5pt}
\par \noindent \textbf{Acknowledgements:} 
This work was in part supported by the Turing 2.0 `Enabling Advanced Autonomy' project funded by the EPSRC and the Alan Turing Institute. 
The research is also supported by the Biome Health Project funded by WWF-UK.

\clearpage
\bibliography{main}

\clearpage
\appendix

\section{Additional Results}

Here we present additional results from experiments and ablation studies that highlight the performance gains and the consistency of SVL-Adapter. 
We maintain the categorisation of datasets, \ie ``Standard'' and ``Challenging'', as defined in the main paper.

\subsection{Better self-supervised features improve adaptation performance} 
\begin{figure}[h]
\centering
\includegraphics[width=1.0\textwidth]{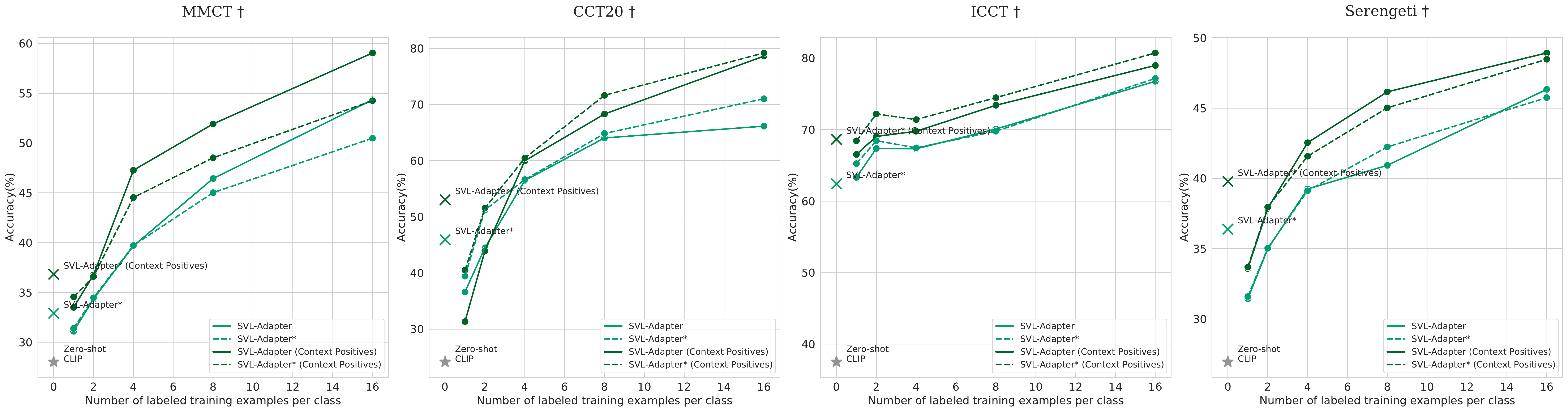}
\vspace{-15pt}
\caption{By using a better self-supervised task (``Context Positives'' from~\cite{Pantazis_2021_ICCV}) to encode features with $E_s$, we achieve consistent gains in zero- and low-shot classification across the four camera trap datasets for both SVL-Adapter and SVL-Adapter*. 
}
\label{fig:avg_results_context_vs_augment}
\end{figure}

In Figs.~3 and 4 of the main paper, we see how the vanilla version of SimCLR \cite{chen2020simple} can help CLIP adapt especially across challenging downstream tasks. Additionally, to support the assumption that the developments in self-supervised learning are orthogonal to our approach, we exploit the metadata that are typically available for real-world tasks to formulate a self-supervised task that uses more informative positive pairs instead of self-augmentations as suggested in~\cite{Pantazis_2021_ICCV}. Specifically, for the four camera trap datasets \cite{Pantazis_2021_ICCV,beery2018recognition,islandconservation2020,swanson2015snapshot}, we perform self-supervised training with context-informed positives and observe significant improvements in both zero- and low-shot classification when the features are exploited by our suggested SVL-Adapter* and SVL-Adapter respectively (Fig.~\ref{fig:avg_results_context_vs_augment}). 
From the results in Fig.~\ref{fig:avg_results_context_vs_augment} we see that the inclusion of context-informed positives instead of self-augmentations during the self-supervised tasks can lead to improvements in both zero-shot and low-shot learning by our proposed  SVL-Adapter* and SVL-Adapter approaches. In addition, we show that a simple Triplet loss can replace SimCLR and remain effective on adaptation (see Table~\ref{tab:ablations_comb}).

\subsection{Better CLIP models translate to better performance}
\begin{figure}[h]
\centering
\includegraphics[width=1.0\textwidth]{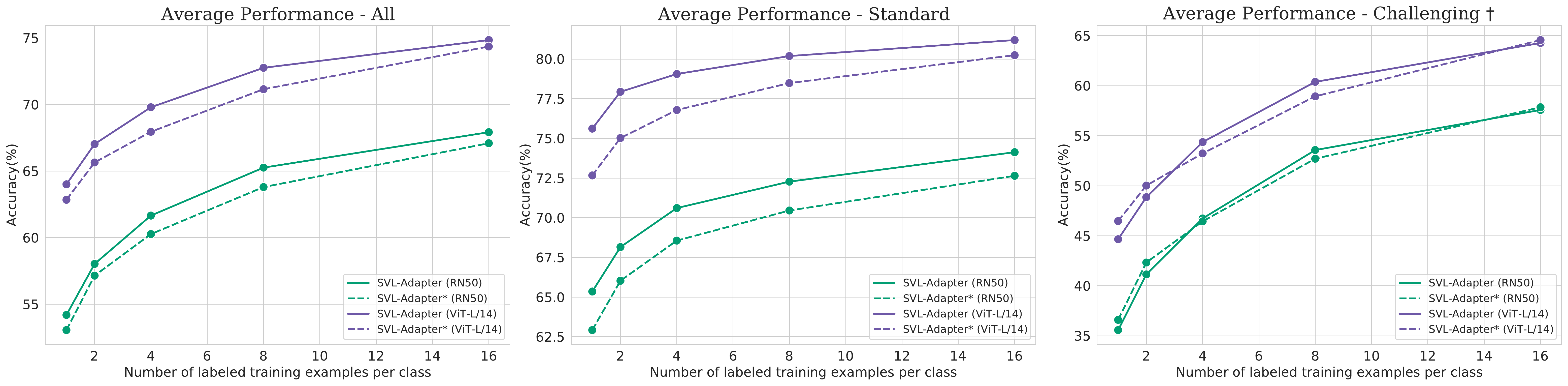}
\vspace{-15pt}
\caption{Low-shot evaluation of SVL-Adapter and SVL-Adapter* fused with Zero-shot CLIP with a ResNet50 backbone as in the main paper versus Zero-shot CLIP with a more advanced large visual transformer. We observe significant gains when fusing the logits that come out of the adapter with a more advanced Zero-shot model (Purple) both in ``Standard'' and ``Challenging'' tasks.}
\label{fig:avg_results_rn50_vs_vitl14_clip}
\end{figure}

The experiments in the main paper use ResNet50 \cite{he2016deep} as the backbone architecture for both CLIP and the self-supervised feature encoder component of SVL-Adapter. Given that the adapter module $A_s$ is expecting the output of the pretrained self-supervised feature encoder $E_s$ we understand than increasing the capacity of its backbone model would make the approach less computationally efficient. 
However, we experiment with the size of the frozen CLIP Encoder given that it is only used for inference and thus it would not be prohibitive for practitioners, even those  with limited resources. Specifically, we replace the ResNet50 CLIP with a ViT/L-14, a large vision transformer~\cite{dosovitskiy2021image}, \ie CLIP's best performing variant according to \cite{radford2021learning}. 
As expected, in Fig.~\ref{fig:avg_results_rn50_vs_vitl14_clip} we observe that the improvements in the CLIP backbone are also reflected in the SVL-Adapter combining them with the adapter features both across ``Standard'' and ``Challenging'' tasks.

\subsection{SSL features alone are not very effective}
\begin{figure}[h]
\centering
\includegraphics[width=1.0\textwidth]{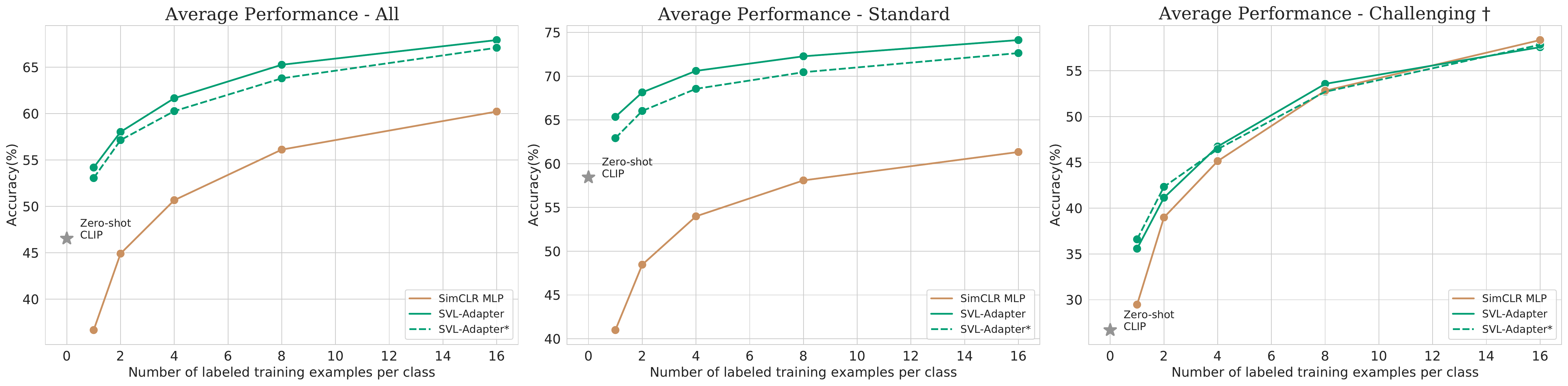}
\vspace{-15pt}
\caption{Here we compare Zero-shot CLIP and adapted self-supervised features (Brown) with our suggested SVL-Adapter variants that essentially are blending the predictions that come out of the two as described in Sec.~3.2.1 of the main paper. We confirm that SVL-Adapter is always better than its constituent parts when used independently. Interestingly, we see the larger benefits of using SVL-Adapter instead of pure self-supervised feature adaptation in the ``Standard'' datasets.}
\label{fig:avg_results_pure_ssl_vs_fusion}
\end{figure}

Even though in the main paper we showed the advantage of using a fusion of adapted self-supervised features and the impressive outputs of Zero-shot CLIP by comparing to existing baselines, we have not compared SVL-Adapter with a variant that only relies on the visual adaptation of the self-supervised features. Here, we compare SVL-Adapter with the adaptation of the SSL features using a similar adapter module but without the blending component in order to understand the necessity of including the CLIP component as described in Eqn.~3 and Sec.~3.2.1 of the main paper. 
As we see in Fig.~\ref{fig:avg_results_pure_ssl_vs_fusion}, SVL-Adapter and SVL-Adapter* outperform Zero-shot CLIP and the independently adapted self-supervised features (SimCLR MLP) on average; thus, a combination, similar to the one proposed with SVL-Adapter in Eqn.~3 in the main paper, should always be preferred as a universal solution.

\subsection{Replacing the encoder with ImageNet or CLIP features}
\begin{figure}[h]
\centering
\includegraphics[width=1.0\textwidth]{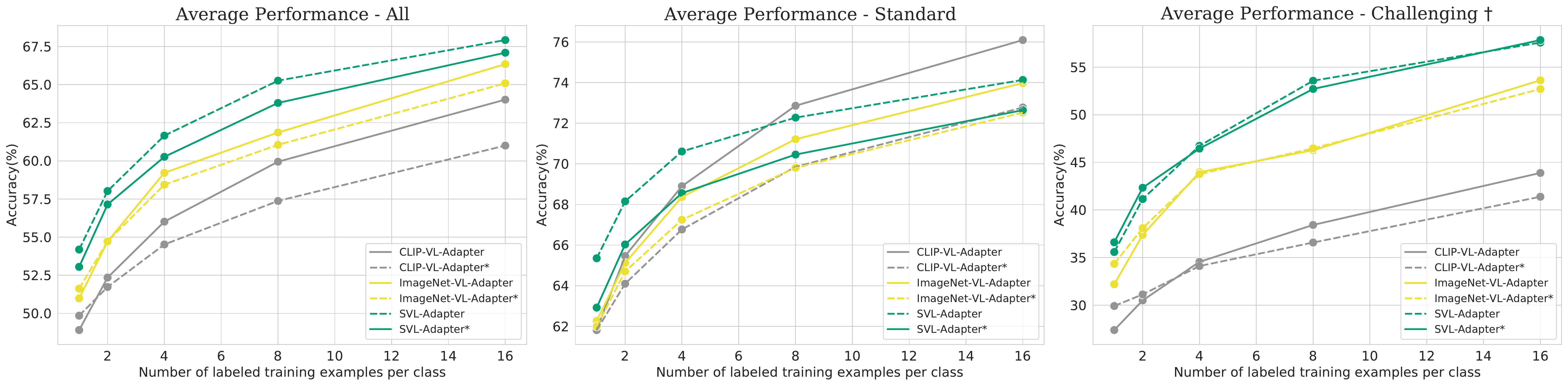}
\vspace{-15pt}
\caption{Here we compare the low-shot adaptation of our approach (SVL-Adapter) with two variants that tune a similar adapter on top of supervised ImageNet features (Yellow) and CLIP features (Grey). We see that adapting from self-supervised features keeps giving the best results on average, especially on challenging tasks while we also observe decent performance from the ImageNet-based adapter.}
\label{fig:avg_results_svl_imagenet_clip}
\end{figure}

To better understand the importance of the features extracted by a self-supervised encoder $E_s$, we replace them with features extracted by either the CLIP ResNet50 features or an ImageNet pretrained ResNet50 while keeping the rest of the components fixed. In Fig.~\ref{fig:avg_results_svl_imagenet_clip} we see that SVL-Adapter keeps delivering the best performance on average and especially on the challenging tasks when compared with a similar adapter that starts from ImageNet or CLIP ResNet50 features. Perhaps surprisingly, we see that adaptation from ImageNet features is clearly superior to adaptation from the CLIP features when evaluated on the challenging tasks. Despite the large size of the corpus that CLIP is pretrained on, it seems that the quantity of data does not necessarily translate to the visual diversity needed to tackle challenging tasks that potentially fall out of distribution, which could be better addressed with ImageNet pretrained models.

\subsection{The impact of $\lambda$ parameter on SSL and CLIP fusion}

\begin{figure}[hb]
\centering
\resizebox{1.0\linewidth}{!}
{
\includegraphics[width=1.0\textwidth]{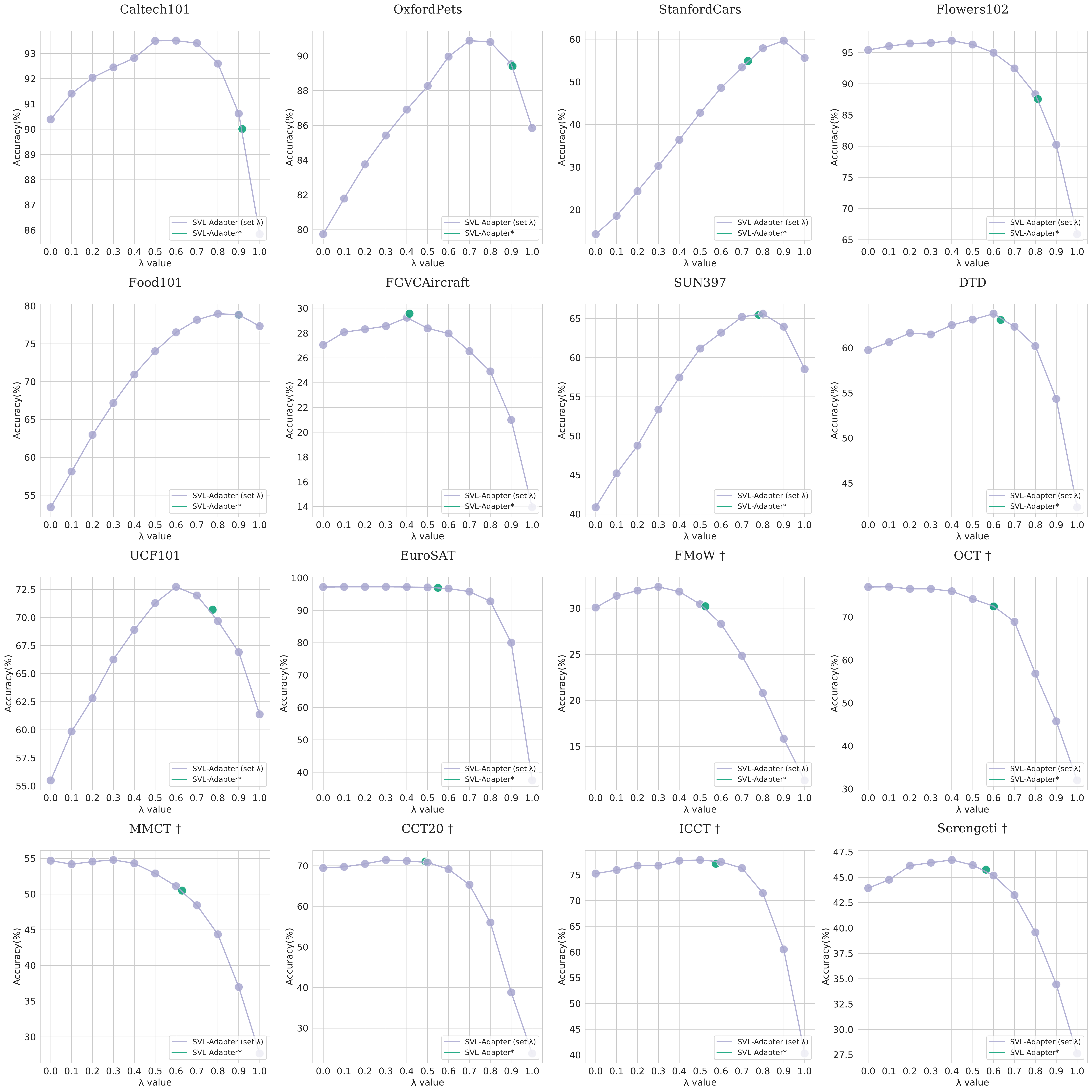}
}
\vspace{-15pt}
\caption{Per-dataset 16-shot, test Top-1 accuracy for the 16 different datasets across various values of the $\lambda$ parameter along with the proposed SVL-Adapter*. In each case, we report the average of three runs. Datasets marked with $\dagger$ are added by us, and pose a significantly greater challenge to existing methods. 
}
\label{fig:main_results_across_lambdas}
\end{figure}

As we have seen in the main text, some tasks are easier for vision-language models that are pre-trained on online image and text pairs while some are quite challenging. Thus, a constant fusion parameter $\lambda$ (see~Eqn.3) between adapted self-supervised and Zero-shot CLIP features for all cases would yield sub-optimal results. We run an ablation experiment on SVL-Adapter where we vary $\lambda$ with values ranging from 0 to 1 and compare with the parameter-free SVL-Adapter* approach proposed in Sec.~3.2.1. Note that a low value of $\lambda$ here means that the self-supervised features adapted via low-shot learning are taken more into account. After observing Fig.~\ref{fig:main_results_across_lambdas}, we can see that the ``Challenging'' tasks ($\dagger$) need a lower value of $\lambda$ to perform well on the test set while most of the other datasets reach peak performance with high $\lambda$ values. In addition, it is shown that the SVL-Adapter* which picks $\lambda$ based on CLIP's average prediction confidence is typically near the top performing points of the curve.

\subsection{Ablations on the components of SVL-Adapter}

We perform multiple ablations on the components of SVL-Adapter, varying the pre-training source, CLIP's vision backbone and the underlying self-supervised method. The results can be found on Table~\ref{tab:ablations_comb}. Given this table, we confirm that fusing self-supervised features learnt for the task at hand with Zero-shot CLIP gives the best performance. Moreover, replacing either CLIP's backbone with a more advanced architecture or the self-supervised technique with a better one can further boost performance. Indicatively, the average performance of Tip-Adapter-F, the best performing baseline on the four camera trap datasets, is 50.8, which is surpassed by a big margin even by a simple Triplet loss trained in a self-supervised manner in the given dataset.

\begin{table}[h]
\centering

\begin{tabular}{|c|c|c|}
\hline
CLIP Visual Encoder  &Adapted Features &Top-1 Accuracy \\ \hline
{-} & SimCLR RN50 &60.2$^\ast$   \\ \hline 
RN50 &{-} &46.5$^\ast$  \\ \hline 
RN50 &ImageNet RN50 &66.3$^\ast$ \\ \hline 
RN50 &CLIP RN50 &64.0$^\ast$ \\ \hline 
RN50 &SimCLR RN50 &{\bf 67.9}$^\ast$ \\ \hline \hline%
ViT-L/14 &{-} &60.1$^\ast$   \\ \hline 
ViT-L/14 &SimCLR RN50 &{\bf 74.9}$^\ast$ \\ \hline \hline%
RN50 & SimCLR RN50 &60.4$^\ddagger$\\ \hline 
RN50 & Triplet RN50 &57.0$^\ddagger$ \\ \hline 
RN50 & SimCLR RN50 + CP~[39] &{\bf 66.4}$^\ddagger$ \\ \hline 
\end{tabular}
\vspace{10pt}
\caption{
Ablation experiments. $^\ast$ are 16-shot Top-1 accuracy for all datasets and  $^\ddagger$ are camera trap datasets only.
`CLIP Visual Encoder' is the zero-shot backbone and `Adapted Features' are the features used by SVL-Adapter's adapter.
(Top) Ablations on the visual representations of SVL-Adapter. 
(Middle) Impact of different CLIP visual encoders. 
(Bottom) Impact of different SSL methods. 
}
\label{tab:ablations_comb}
\end{table}

\subsection{Understanding why SSL helps}

\begin{figure}[t]
\centering
\includegraphics[width=1.0\textwidth]{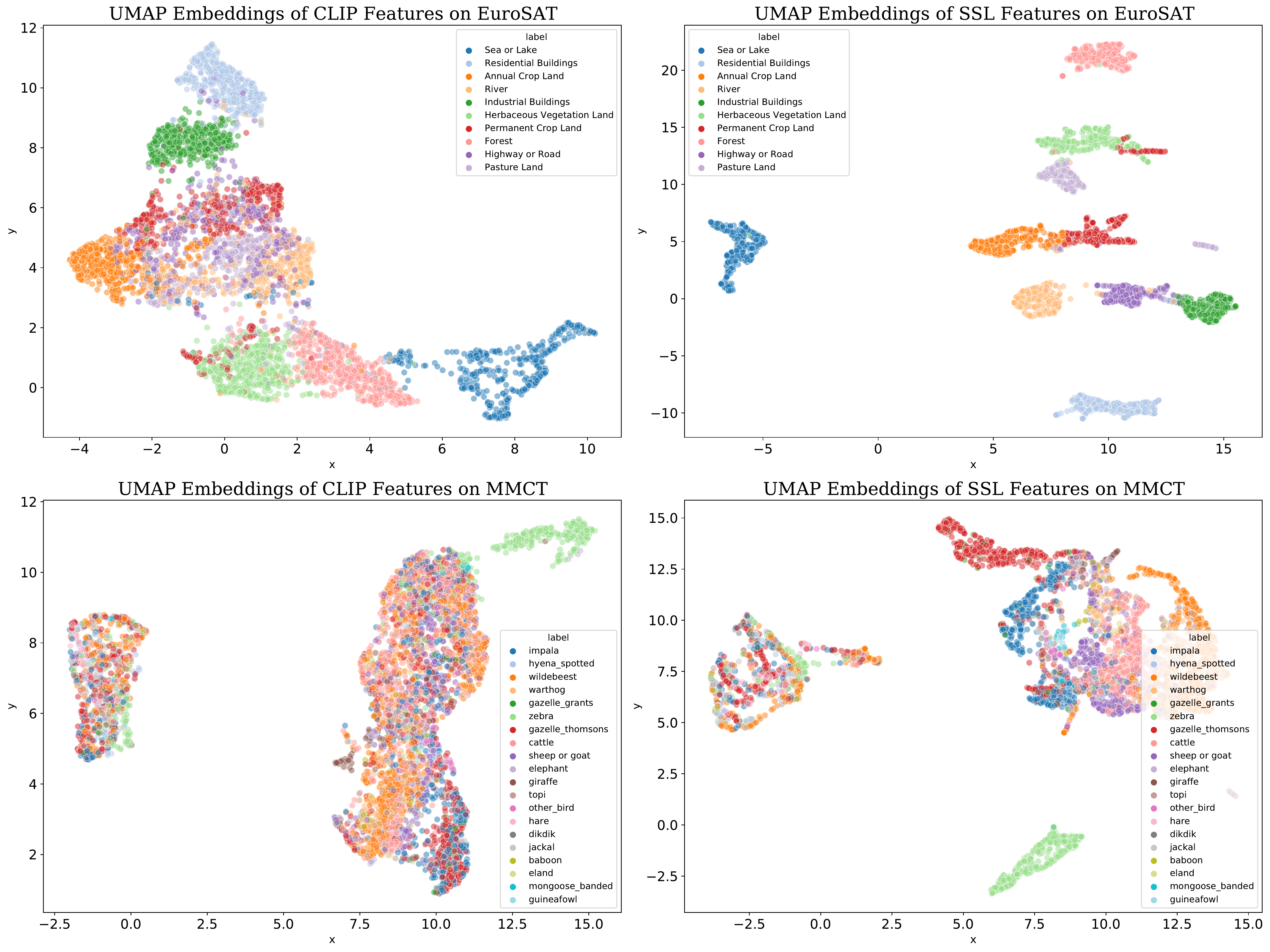}
\vspace{-15pt}
\caption{2-dimensional UMAP embeddings of test image features coming from CLIP (Left) or Self-Supervised (Right) ResNet50 feature extractors. For both EuroSAT (top) and MMCT (bottom) we observe that the self-supervised learning features correspond to a better separation between the dataset classes.
}
\label{fig:umap_features_ssl_vs_clip}
\end{figure}

To try to understand why adaptation from self-supervised features is superior on the challenging tasks, we plot their 2-dimensional embeddings on test images along with the respective projections of CLIP features. We use UMAP~\cite{mcinnes2018umap} for the dimensionality reduction across EuroSAT and MMCT. As we  observe from Fig.~\ref{fig:umap_features_ssl_vs_clip}, the embeddings of the self-supervised learning features (right) provide a better separation of the underlying classes in the unseen test set, while the 2-dimension projection of the CLIP extracted features is not as well structured and overlaps in many cases.

\subsection{Quantitative results in tabular form}
For a more detailed view, we present the quantitative results from Fig.~3 of the main paper in a tabular format (Table~\ref{tab:res_comb}).

\begin{table}[h]

\centering

\begin{tabular}{|l|c|c|c|c|}
\hline
\multicolumn{5}{|c|}{\bf Average Performance - All}\\ \hline 
     & CoOp & CLIP-Adapter  & Tip-Adapter-F   & SVL-Adapter \\ \hline \hline
1-shot  &48.1 &49.7 &53.6  & {\bf 54.2} \\ \hline
2-shot  &50.9 &53.6 &55.2  & {\bf 58.0} \\ \hline
4-shot  &54.4 &56.8 &58.0  & {\bf 61.7} \\ \hline
8-shot  &59.5 &59.9 &61.6  & {\bf 65.3} \\ \hline
16-shot &63.9 &64.0 &66.2 & {\bf 67.9} \\ \hline 
\end{tabular}

\vspace{5pt}
\begin{tabular}{|l|c|c|c|c|}
\hline
\multicolumn{5}{|c|}{\bf Average Performance - Standard}\\ \hline 
     & CoOp & CLIP-Adapter  & Tip-Adapter-F   & SVL-Adapter \\ \hline \hline
1-shot  &59.8 &62.9 &64.9  & {\bf 65.4} \\ \hline
2-shot  &62.8 &66.0 &67.2  & {\bf 68.2} \\ \hline
4-shot  &67.4 &69.3 &70.4  & {\bf 70.6} \\ \hline
8-shot  &70.8 &72.3 & {\bf 73.3}  &72.3 \\ \hline
16-shot &74.5 &75.5 & {\bf 76.9} &74.1 \\ \hline 
\end{tabular}

\vspace{5pt}
\begin{tabular}{|l|c|c|c|c|}
\hline
\multicolumn{5}{|c|}{\bf Average Performance -  Challenging $\dagger$}\\ \hline 
     & CoOp & CLIP-Adapter  & Tip-Adapter-F   & SVL-Adapter \\ \hline \hline
1-shot  &28.5 &27.8 &34.7  & {\bf 35.6} \\ \hline
2-shot  &31.2 &33.0 &35.3  & {\bf 41.1} \\ \hline
4-shot  &32.7 &35.9 &37.5  & {\bf 46.7} \\ \hline
8-shot  &40.8 &39.4 &42.1  & {\bf 53.6} \\ \hline
16-shot &46.3 &44.8 &48.3 & {\bf 57.6} \\ \hline 
\end{tabular}

\vspace{10pt}

\caption{Top-1 Accuracy from Fig.~3 of the main paper, but in tabular format. (Top) Averaged across all 16 datasets with the corresponding \emph{zero-shot} results for CLIP and our zero-shot SVL-Adapter* being 46.5\% and 52.5\% respectively.
(Middle) Averaged across the ``Standard'' datasets with the corresponding \emph{zero-shot} results for CLIP and our zero-shot SVL-Adapter* being 58.4\% and 63.1\% respectively.
(Bottom) Averaged across the ``Challenging'' datasets with the corresponding \emph{zero-shot} results for CLIP and our zero-shot SVL-Adapter* being 26.7\% and 34.8\% respectively.}

\label{tab:res_comb}
\end{table}

\section{Implementation Details}

\subsection{Prompt templates}

Across all experiments that involved $E_t$, \ie the text encoder of CLIP , we used a prompt template for each of the datasets. As mentioned in the main text, for the ``Standard'' datasets we use the simple prompts suggested by \cite{radford2021learning} and adopted by the subsequent vision-language adaptation works~\cite{zhou2021learning,gao2021clip,zhang2021tip}. With respect to the ``Challenging'' tasks, we utilized the generic prompt "a photo of a \{label\}." for MMCT, CCT20, ICCT, Serengeti, and FMoW and "an OCT scan of \{label\} retina." for the Optical Coherence Tomography (OCT) dataset. To make sure our class names are suitable for the text encoder of CLIP, we transformed some of them in a human-friendly format. For example, in OCT we replaced the disease names "CNV" and "DME" with their full names "Choroidal Neovascularization" and "Diabetic Macular Edema" respectively. In addition, we replace underscores with blanks and separate class names (\eg from "gazelleGrants" to "gazelle grants") where necessary.

\subsection{Training details}
In addition to the implementation details provided in Sec.~4.1 of the main paper, we present more details about our modeling choices.

\noindent{\bf Self-supervised pretraining.} For self-supervised learning we train for 200 epochs with batch size 256 and learning rate 0.03 and cosine annealing. For optimization we use Stochastic Gradient Descent (SGD) with 0.9 momentum and weight decay 0.0005. On top of the ResNet 50 backbone, we used a projection neural network with 512 hidden size and 128 embedding size. For the image transformations on the $112 \times 112$ sized images of the self-supervised task we use
cropping with random resizing, random horizontal flipping, color jittering and grayscale conversion. For the context positive mining approach used to generate Fig.~\ref{fig:avg_results_context_vs_augment} we used the metadata and implementation details as suggested by \cite{Pantazis_2021_ICCV}.

\noindent{\bf Adapter tuning.} The downstream task adaptation takes place on top of the frozen features generated by the feature encoders, \eg by the ResNet50 encoder $E_s$ that comes from self-supervised pretraining. We train the adapter module $A_s$ of SVL-Adapter for 50 epochs with batch size of 32. Specifically, this module corresponds to a 2-layer MLP with a 256 hidden layer that receives the frozen features from $E_s$ andf outputs the class logits which are used in the fusion stage described in~Eqn.3 of the main paper. For optimization, we utilized Adam with learning rate 0.001 and a cross entropy loss between the logits and the few-shot sample labels or pseudolabels in SVL-Adapter and Zero-shot SVL-Adapter respectively. For the scenario where we use a small validation set to tune the $\lambda$ parameter (Sec.~3.2.1 of main paper), we use 1 and 2 samples for 1-shot and 2-shot classification respectively and 4 samples for the rest of the labeled 4-, 8-, and 16-shots, similarly with preceding works \cite{gao2021clip,zhang2021tip}.

\end{document}